# Color and Shape Content Based Image Classification using RBF Network and PSO Technique: A Survey

Abhishek Pandey
Dept. of CSE
UIT-RGPV Bhopal, India
pandeyabhishek14496@gmail.com

Prof. Anjna Jayant Deen
Dept. of CSE
UIT-RGPV Bhopal , India
anjnadeen@yahoo.com

Dr. Rajeev Pandey
Dept. of CSE
UIT-RGPV Bhopal , India
rajeev98iet@gmail.com

**Abstract**

**The improvement of the accuracy of image query retrieval used image classification technique. Image classification is well known technique of supervised learning. The improved method of image classification increases the working efficiency of image query retrieval. For the improvements of classification technique we used RBF neural network function for better prediction of feature used in image retrieval.Colour content is represented by pixel values in image classification using radial base function(RBF) technique. This approach provides better result compare to SVM technique in image representation.Image is represented by matrix though RBF using pixel values of colour intensity of image. Firstly we using RGB colour model. In this colour model we use red, green and blue colour intensity values in matrix.SVM with partical swarm optimization for image classification is implemented in content of images which provide better Results based on the proposed approach are found encouraging in terms of color image classification accuracy.**

Introduction

classifying Image classification is defined as the task of the number of images into (semantic) categories based on the available supervised data. The main objective of digital image classification procedure is to categorize the pixels in an image into land over cover classes. The output is thematic image with a limited number of feature classes as opposed to a continuous image with varying shades of gray or varying colors representing a continuous range of spectral reflectance [20].
RBF function is a neural network approach. It is based on function values which is measure by origin. The distance show colour intensity of image. Image features are colour, texture, shape and size. Large collections of images are becoming available to the public, from photocollection to web pages or even video databases. Sincevisual media requires large amounts of memory andcomputing power for processing and storage, there is aneed to efficiently index and retrieve visual informationfrom image database [21].
The idea of RBFderives from the theory of function approximation. We have already seen how Multi-Layer Perception (MLP) networkswith a hidden layer of sigmoid units can learn to approximate functions. RBF Networks take a slightly different approach [11, 17]. Their main features are: They are two-layer feed-forward networks. The hidden nodes implement a set of radial basis functions (e.g. Gaussian functions). The output nodes implement linear summation functions as in an MLP. The network training is divided into two stages: first the weights from the input tohidden layer are determined, and then the weights from the hidden to output layer [16].
The Image classification using SVM classifier,RBF classifier and POS optimization technique based on content of image are providing comparatively result. Efficient indexing and Extraction of large number of color images, classification plays an important and challenging role. The main aim of this research work is devoted to finding suitable representation for images and classification generally requires comparison of images depending on the certain useful features [5].

Literature Survey

(1) Xiao-Qing Shang, Guo-Xiang Song, Biao Hou in China in 2003. They carried out work on "content based texture image classification." A new method for content based texture image classification is proposed using support vector machine of the image, which combines the characteristics of Brushlet and Wavelet transform.

(2) P. J.Gri ffi Ths, J. S. Marsland and W.Eccleston in Leverpoolin 2003.They discussed work on "A Study of Noise in PWM Neural Networks". This paper shows the effect of introducing noise to the weight set and at the input to the neuron. The MLP investigated is tolerant to noise added at the input to the neuron and therefore could be implemented using the PWM neural network with the RC time constant set close to the PWM period.

(3) Keiji Yanai in Japan in 2003.They work out on "Generic Image Classification Using Visual Knowledge on the Web." In this paper, They describe a generic image classification system with an automatic knowledge acquisition mechanism from the World Wide Web.

(4) Luca Piras and Giorgio Jiacinto in University of Cagliari in Italy in 2010. They proposed work on "Unbalanced learning in content-based image classification and retrieval."In this paper we propose a technique aimed at artificially increasing the number of examples in the training set in order to improve the learning capabilities, reducing the unbalance between the semantic class of interest, and all other images. The proposed approach is tailored to classification and relevance feedback techniques based on the Nearest-Neighbor paradigm. A number of new points in the feature space are created based on the available training patterns; so that they better represent the distribution of the semantic class of interest.

(5) Saurabh Agrawal, Nishchal K Verma, Prateek Tamrakar, Pradip Sircar in Indian Institute ofTechnology Kanpur, India at 2011.They work on "Content Based Color Image Classification using SVM."They implement classification of image using SVM classifier in the colour content of image. They use optimal hyper planes technique thorough support vector machine.In this paper, they use color image classification on features extracted from histograms of color components. The benefit of using color image histograms are better efficiency, and insensitivity to small changes in camera view-point i.e. translation and rotation.

(6) Siu-Yeung Cho inthe University of Nottingham Ningbo China in 2011.They research on"Content Based Structural Recognition for Flower Image Classification." In this paper, a study was made on a development of content based image retrieval system to characterize flower images efficiently. In this system, a method of structural pattern recognition based on probabilistic based recursive model is proposed to classify flower images.

(7) Giuseppe Amato, Fabrizio Falchi and Claudio Gennaro in Pisa, Italy in 2011.They carried out work on"Geometric consistency checks for kNN based image classification relying on local features." In this paper In this paper we propose a technique that allows one to use access methods for similarity searching, such as those exploiting metric space properties, in order to performKNN classification with geometric consistency checks.

(8) Wang Xing Yuan, Chen Zhi feng and Yunjiao Jiao in China in 2011.They carried out work on "An effective method for colour image retrieval based on texture." They proposed thatan effective colour image retrieval method based on texture, which uses the colour occurrence matrix to extract the texture feature and measure the similarity of two colour images.

(9) Yu Zeng , Jixian Zhang , J.L. Van Genderen , Guangliang Wang  Chinese Academy of Surveying and Mapping, Beijing , P.R.China in 2012. They research on"SVM-based Multi-textural Image Classification and Its Uncertainty Analysis."This paper presents a supervised image classification method which ismultiple and multi-scale texture features and support vector machines (SVM).

(10)Masato Yonekawa and Hiroaki Kurokawa in the School of Computer Science, Tokyo University of Technology, Tokyo Japan in 2012.They proposed on "The Content-Based Image Retrieval using the Pulse Coupled Neural Network." In this paper they proposed a learning method to define the parameters in the PCNN for image matching. The learning method improves the performance of image matching using thePCNN. On the other hand, recently, a lot of researches on the Content-Based Image Retrieval (CBIR) have been studied.

(11) Methaq Gaata ,Sattar Sadkhn, Saad Hasson Montpellier, France in 2012.They work on"Reference Quality Metric Based on Neural Network for Subjective Image WatermarkingEvaluation."In this work the goal of IQA research is to design computational models which have ability to provide an automatic and efficient way to predict visual quality in a way that is consistent with subjective human evaluation.

(12) R. Venkata Ramana Chary, D. Rajya Lakshmi and K.V.N. Sunitha Tiruvannamalai, TN., India In December, 2012 carried out work on "Image

Searching Based on Image Mean Distance Method." They discussed that when the size of database is increasing image similarity finding .It is big task for the researchers to give the efficient solutions. Content-Based Image Retrieval (CBIR) systems are used in order to retrieve image from image dataset. In our proposed method, we are utilizing clustering method for retrieve the images.

(13) Mihir Jain,Rachid Benmokhtar,Patrick GrosINRIA Rennes in 2012.They carried out work on "Hamming Embedding Similarity-based Image Classification." They propose a novel image classification frame- work based on patch matching. More precisely, we adapt the Hamming Embedding technique, first introduced for image search to improve the bag-of-words representation. We propose a mapping method to cast the scores output by the Hamming technique into a proper similarity space.

(14) Amel Znaidia, Aymen Shabou, Adrian Popescu, Hervé le Borgne , Céline Hudelot in france in 2012.They carried out work on "Multimodal Feature Generation Framework for Semantic Image". Classification unified framework which mixes textual and visual information in a seamless manner.

(15)Feilong Cao,Bo liu and Dong Sun Park in china in 2012. They research on "Image classification based on effective extreme learning machine." In this work, a new image classification method is proposed based on extreme k means (EKM) and effective extreme learning machine. The proposed processes has image decomposition with curve let transform, reduces dimensiolity with discriminative locality alignment (DLA).

(16)Yan leng, Xinyan Xu and Guanghui Qi in china in 2013. They carried out work on "Combining active learning and semi supervised learning to construct SVM classifier." In this work active semi supervised SVM algorithm perform much better than the pure SVM algorithm

(17) Ming Hui Cheng,Kao Shing Hwang Jyh Horng Jeng and Nai Wei lin Taiwan in 2013. They work on "classification based video super resolution using artificial neural networks." In this study, they proposed to enhance low resolution to high resolution frames. The proposed method consists of four main steps classification motion trace volume collection temporal adjustment and ANN prediction classifier is designed based on the edge properties of a pixel in the frame to identify the spatial information.

(18) Marko Tkalcic , AnteOdic, Andrej Kosir and Jurij Tasic member of IEEE in feb 2013.They carried out work on " Affective Labeling in a Content-Based Recommender System for Images."In this paper we present a methodology for the implicit acquisition of affective labels for images.It is based on an cotent detection technique that takes as input the video sequences of users facial expressions. It extracts Gabor low level features from the video frames and employs a k nearest neighbor's machine learning technique to generate affective labels in the valence-arousal-dominance space.

(19) Serdar Arslan, Adnan Yazici, Ahmet Sacan,Ismail H,Toroslu Esra Acar in USA in 2013.They proposed work on "Comparison of feature based and image registration based retrieval of image data using multidimensional data access methods" They proposed that multidimensional scaling can achieve comparable accuracy while speeding up the query times significantly by allowing the use of spatial access methods.

**Comparison between RBF network and other classification technique:**

(1)The classification techniques are not providing better optimal result. Some techniques are traditional. Radial basis function network technique is artificial neural network technique. It is provide optical classification which is based on supervised learning and training data set.

(2) As a SVM classifier SVM suffering two problems:
    (i) How to choose optimal feature sub set input.
    (ii) How to set best kernel parameters.

These problems influence the performance and accuracy of support vector machine. Now the pre-sampling of feature reduced the feature selection process of support vector machine for image classification.

(3) For the improvements of classification technique we used RBF neural network function for better prediction of feature used in image retrieval. Our proposed method optimized the feature selection process and finally sends data to multiclass classifier for classification of data.

## Conclusion

After survey of papers we find that Image is classified through its content like colour, texture, shape and size. In this paper, feature extraction of image is based on colour, shape and size content. Feature extraction of image is optimal. Optimal Feature of image is classified by RBF classifier. Classification of image is using RBF neural network. Radial basis network (RBF) is a artificial neural network technique. It is provide supervised classification of image features. The Gaussian distributionfunction is used in hidden unit of RBF network. Classifications of optimal feature of image are implemented by RBF algorithm.In Radial basis function, feature value is represented in matrix form. In this technique distance of pixel is measured optically with origin. This technique provides better performance and accuracy of image compare to KNN and SVM classification approach.For the improvement of support vector machine classifier we used RBF neural network and POS optimization technique. Our empirical result shows better efficiency instead of support vector machine classifier.This approach provides better result of colour feature of image classification.

## Future scope of work

Radial basis function network have a hidden processing unit where apply different type of training algorithm. It increases image quality through modification of algorithm. Timing of classification can also improve compare to other classification technique. RBF network can be applied to other type of classification technique of image processing. The Accuracy of classification of image play vital role in medical field. RBF network and POS optimization technique usingtrained feature so this technique can be more enhancement are possible in the future.